\documentclass{sig-alternate}
\usepackage{color}
\usepackage[numbers,sort&compress]{natbib}
\usepackage{graphicx}
\usepackage{epstopdf}

\usepackage{fancyhdr}
 
\usepackage[linesnumbered,ruled]{algorithm2e}

\SetCommentSty{mycommfont}
\newcommand{\ignore}[1]{}
\begin{document}

\title{Runtime Configurable Deep Neural Networks for
           Energy-Accuracy Trade-off}

\author{Hokchhay Tann, Soheil Hashemi, R. Iris Bahar, Sherief Reda \\
      \affaddr{School of Engineering}
      \affaddr{Brown University}
      \affaddr{Providence, RI 02912}\\
      \email{\large \{hokchhay\_tann, soheil\_hashemi, iris\_bahar, sherief\_reda\}@brown.edu}
}

\ignore{
\numberofauthors{4}
\author{
\alignauthor
Hokchhay Tann\\
      \affaddr{School of Engineering}\\
      \affaddr{Brown University}\\
      \affaddr{Providence, RI 02912}\\
      \email{\large hokchhay\_tann@brown.edu}
\alignauthor
Soheil Hashemi\\
      \affaddr{School of Engineering}\\
      \affaddr{Brown University}\\
      \affaddr{Providence, RI 02912}\\
      \email{\large soheil\_hashemi@brown.edu}
\and
\alignauthor
R. Iris Bahar\\
      \affaddr{School of Engineering}\\
      \affaddr{Brown University}\\
      \affaddr{Providence, RI 02912}\\
      \email{\large iris\_bahar@brown.edu}
\alignauthor 
Sherief Reda\\
      \affaddr{School of Engineering}\\
      \affaddr{Brown University}\\
      \affaddr{Providence, RI 02912}\\
      \email{\large sherief\_reda@brown.edu}
}
}


\maketitle
\begin{abstract}
We present a novel dynamic configuration technique for deep neural networks that permits step-wise energy-accuracy trade-offs during runtime. Our configuration technique adjusts the number of channels in the network dynamically depending on  response time, power, and accuracy targets. To enable this dynamic configuration technique, we co-design a new training algorithm, where the network is incrementally trained such that the weights in channels trained in earlier steps are fixed. Our technique provides the flexibility of multiple networks while storing and utilizing one set of weights.
We evaluate our techniques using both an ASIC-based hardware accelerator as well as a low-power embedded GPGPU and show that our approach leads to only a small or negligible loss in the final network accuracy.
We analyze the performance of our proposed methodology using three well-known networks for MNIST, CIFAR-10, and SVHN datasets, and we show that we are able to achieve up to 95\% energy reduction with less than 1\% accuracy loss across the three benchmarks.
In addition, compared to prior work on dynamic network reconfiguration, we show that our approach leads to approximately 50\% savings in storage requirements, while achieving similar accuracy.
\end{abstract}


\section{Introduction}
\label{sec:Introduction}
Deep neural networks (DNNs) have recently been the focus of many works in the machine learning and computer vision communities~\cite{Farabet:NeuFlow,convnet,svhn}. Thanks to the increase in computational capabilities and availability of big data, large DNNs can be trained in relatively short periods of time allowing them to surpass the performances of other methods or even humans. As a result, many of the current real world image and voice recognition applications such as Google image search~\cite{google}, and Siri voice recognition~\cite{siri} utilize DNNs. The application of such networks in many state-of-the-art recognition and classification problems has highlighted the importance of high throughput and power efficient hardware platforms. While delivering the state of the art results in terms of accuracy, DNNs are extremely demanding in terms of both computation and memory requirements. Such high demands have limited the application of these networks to high-end GPUs, which deliver the required throughput, albeit at a high energy cost.

On the other hand, energy efficiency has emerged as one of the most significant concerns in the  computer architecture community. DNNs have high memory and computational  power demands, which translate directly to high power consumption. As a result, power and memory efficient implementations of DNNs that can achieve high throughput in an energy-efficient manner are of great importance.

DNNs have an inherent error tolerance that originates from both their deployed applications as well as their very nature, which can compensate for errors in the training process. This error tolerance can be exploited by approximate computing techniques to trade small amounts in accuracy for significant savings in power consumption, silicon area and design complexity.

In this work we propose a novel dynamic configuration methodology that enables DNNs to (1) save energy during runtime without compromising their accuracy, and to (2) meet hard constraints in response time and power consumption with graceful reduction in accuracy. Our contributions are as follows.

\vspace*{-2mm}
\begin{itemize}
\itemsep -.02in
\item To enable dynamic configuration, we propose to adjust the number of active channels per layer of the DNN during runtime. Our technique allows DNNs to be partially or fully deployed, which leads to energy savings and enables the DNN to track runtime constraints. 
\item To enable dynamic configuration without compromising accuracy, we co-design an incremental training algorithm that takes advantage of intrinsic features of neural networks, where we initially train a subset of channels in each layer and gradually add in more channels, while keeping the earlier trained channels fixed. We propose novel methods to adjust the weights to ensure that the network retains the accuracy of the original network when it is fully deployed. Our method offers the flexibility that would arise from using multiple DNNs of different capacities, while only requiring the memory and hardware real estate of one network.

\item We analyze the energy-accuracy trade-offs enabled from our approach in two different scenarios. The first scenario dynamically configures the DNN based on constraints arising during runtime, such as response time, power and energy. The second scenario dynamically adjusts the DNN to save energy as long as the accuracy of the classification results are not compromised. We develop a method to determine the appropriate network size and dynamically adjust it if the wrong inference is detected.

\item We implement and evaluate our proposed methods in two different platforms commonly used within embedded systems: a custom ASIC-based hardware accelerator design and a low-power embedded GPU. For the ASIC implementation, we use an industrial-strength flow in 65 nm technology, and use the flow to evaluate the runtime, power and energy consumption of the hardware. 
\item Using the two platforms, we evaluate our methodology on three well-recognized and diverse classification testbenches using three different network architectures. The three testbenches are MNIST, CIFAR-10 and SVHN datasets \cite{mnist, cifar10, svhn} running on LeNet, ALEXnet and ConvNet respectively \cite{lenet, cifar10, convnet}. Our results show up to 95\% reduction in runtime, with small or no accuracy loss, and with less memory overhead. Further, we provide a systematic method for achieving such savings.
\end{itemize}

The rest of the paper is organized as follows. First, in Section~\ref{sec:PreviousWork}, we provide a short introduction into DNNs and briefly report on recent work targeting low-power and approximate hardware implementations of DNNs. Section~\ref{sec:Methodology} describes our incremental training methodolgy followed by Section~\ref{sec:Runtime}, which describes our opportunistic and constraint-based frameworks. Next, Section~\ref{sec:Results} summarizes our results on the accuracy, runtime benefits, and power benefits. Here, we also include our hardware accelerator design and its characteristics. Finally, Section~\ref{sec:conclusion} summarizes the main contributions of this paper.

\section{Background}
\label{sec:PreviousWork}
\begin{figure}[t]
   \begin{center}
    \includegraphics[width=0.47	\textwidth,natwidth=100,natheight=50]{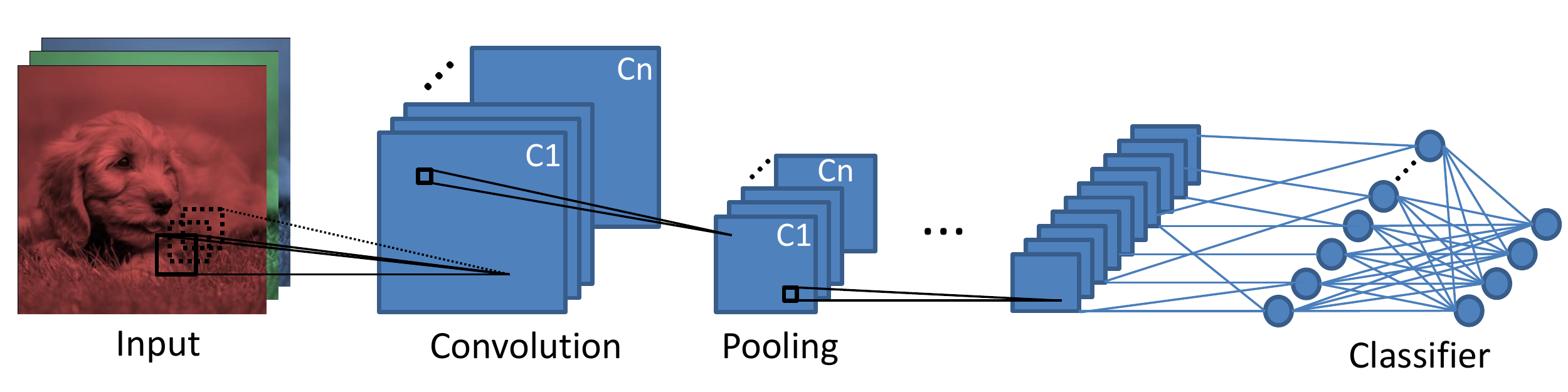}
    \vspace{-0.18in}
    \caption{The structure of a typical DNN.}
    \label{fig:net}
    \end{center}
    \vspace{-0.2in}
\end{figure}

DNNs were originally inspired by the behavior of the human brain. With the availability of increased computational power and large training data, these DNNs have been recently proven to produce state-of-the-art solutions for some of the most challenging problems in computer vision, such as image classification. Typical neural networks consist of several layers, where each layer gets its input from the previous layer and feeds its output only to the next layer. Here, the intermediate values between different layers are named feature maps. Figure~\ref{fig:net} shows the organization and connections of layers in a typical network. In this approach, each layer consists of a number channels, where each channel is responsible for implementing a specific feature map filter. Channels within a layer share the inputs from the previous layer while each using a different set of weights.

\begin{figure}[t]
   \begin{center}
    \includegraphics[scale=0.2]{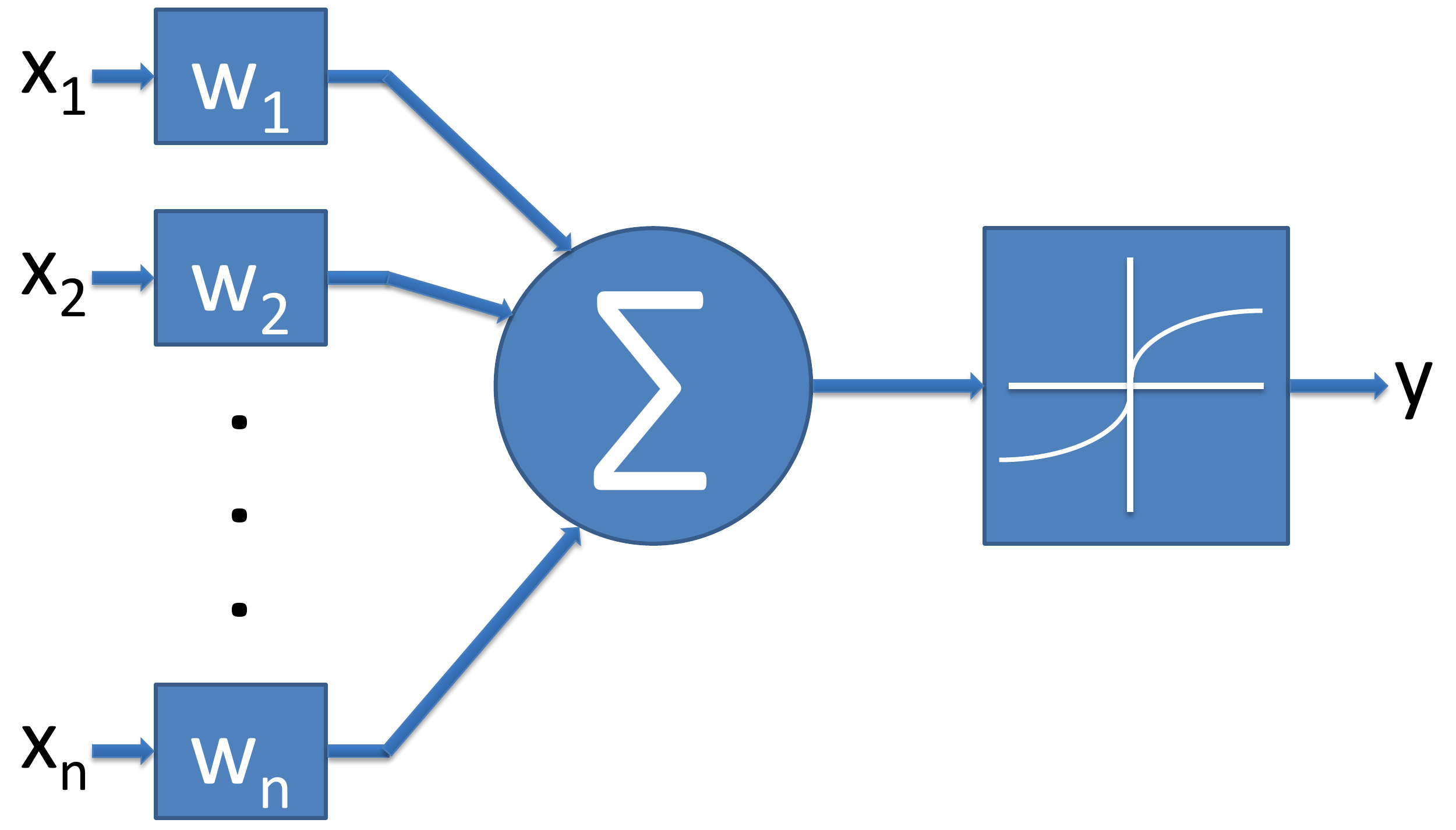}
    \vspace{-0.1in}
    \caption{The structure of a neuron.}
    \label{fig:neuron}
    \end{center}
    \vspace{-0.3in}
\end{figure}

While there is a broad range of different layers available in the literature, three types of layers are most commonly used in DNNs. First, \textit{convolutional layers} have multiple filters, where each filter applies a convolution to the input feature maps. In other words, the convolutional layer performs a weighted sum on a region of the input features. The number of filters directly translates into the number of channels in the respective layer. The building block units of convolutional layers are neurons. Figure~\ref{fig:neuron} shows this functionality of each neuron.

Another class of layers that is closely related to convolutional layers includes \textit{ fully connected layers}. In this layer, each neuron has weighted synaptic connections to all neurons in the previous layer. In other words, a fully connected layer treats its input as a 1-dimensional vector and generates a 1-dimensional vector as the result. Some conventions refer to this layer as simply a convolution layer with $1\times 1$ kernels.

Finally, the third class of layers commonly used in DNNs includes \textit{pooling layers}. These layers extract local information in each feature map by sampling input feature maps. Divided into two main categories (namely, \textit{average pooling} and \textit{max pooling}), these layers are commonly used to reduce the dimensionality of the feature maps and provide translation invariance. Each of these pooling layers can be complemented with a nonlinearity function to add nonlinearity to the system, which has been shown to improve the performance significantly.

DNNs are typically trained using the backpropagation algorithm, where the inference error in the output layer is propagated backward in the form of partial gradients. Each weight and bias is then updated using stochastic gradient descent or one of its variants. After the network is trained, it can be utilized in feedforward mode to evaluate the output results for each input.

Recent interest in DNNs has motivated a broad exploration of viable hardware and software solutions. Many works have focused on optimization of neural networks for effective implementations targeting both FPGAs~\cite{farabet:CNP,Gokhale,Farabet:2009} and custom hardware accelerators~\cite{Kim,Temam,Farabet:NeuFlow}. Other works have focused on optimization of the computation blocks \cite{chakradhar,farabet:CNP,sankaradas}. For example, Farabet {\it et al.}~\cite{farabet:CNP} propose the use of one hardware convolutional operator for implementing  the filtering computation while the rest of the computation is done in software. Different parallelism and locality opportunities are also explored in recent work~\cite{sankaradas,chakradhar,cadambi}. As an example, Chakradhar {\it et al.}~\cite{chakradhar} take advantage of inter-output and intra-output parallelism and design a dynamically configurable hardware design for the forward phase. Zhang {\it et al.}~\cite{zhang} use a rooftop model to identify the best solution given a specific set of resources, thereby mitigating the under-utilization of memory bandwidth and computational logic. Their proposed tile-based custom design can achieve up to 61.62 GOPS using floating point arithmetic. A tile-based hardware accelerator that uses custom-designed memory structures to exploit data locality is proposed in DianNao~\cite{chen} and is capable of performing 452 GOP per second. For our implementation, we exploit a tile-based hardware closely related to the hardware proposed in DianNao.

To simplify the hardware design of DNNs, a number of recent works advocate the use of approximating computing techniques. Du {\it et al.}~\cite{du} propose the use of an approximate multiplier design for weight and input multiplication and conduct a broad design space exploration to determine the best network designs. Venkataramani {\it et al.} propose a methodology in which less sensitive neurons are approximated with precision scaling~\cite{venkataramani}. The power and accuracy results are then evaluated on a customized quality configurable neuromorphic processing engine to report the benefits. In a similar approach, Zhang {\it et al.} propose to remove the less critical neurons in favor of energy reduction~\cite{zhang}. While achieving savings with small accuracy degradation, the benefits from these methods are limited since they only target individual neurons. In addition, the final design is rigid with no runtime reconfigurability. Our proposed incremental training method eliminates both of these constraints without requiring special modification to the network.

Park {\it et al.}~\cite{park} proposed a ``Big/Little'' implementation, where two networks are trained and used to reduce energy requirements. For each input, the little network is first evaluated and the big network is triggered only if the result of the little network is not deemed confident enough. While this work is the closest to ours, our proposed approach differs in that we do not need to store different sets of weights to implement networks with different sizes. This is a significant improvement as DNNs require substantial memory capacity and these memory requirements translate directly to memory transfers and computations. In our approach, intermediate results can be stored for partial use in the bigger network, therefore reducing data storage, transfer, and computation. In addition, we provide a comprehensive methodology to evaluate an application beforehand and identify the best set of configurations such as number of increments as well as the portion of network used in each increment.

In the next section, we present our proposed incremental training and testing methodology, driven by power consumption and memory requirement considerations.

\section{Methodology}
\label{sec:Methodology}

Typical DNN architectures consist of series of convolutions, pooling, and non-linearity. Each convolution layer has varying numbers of channels, each of which is connected to all channels in the layers in front and behind. While the channels contribute to the feature pools for that layer, each channel comes at a cost in terms of weight storage, communication, and computation in a forward pass. In our work, we assume that the number and types of layers and the channels within our testbench architectures are optimal with respect to targeted accuracy. However, we propose to leverage the number of active channels in each layer during runtime to yield energy saving.

\begin{figure}[t]
   \begin{center}
    \includegraphics[scale=0.37,trim=0 0 0 0,clip]{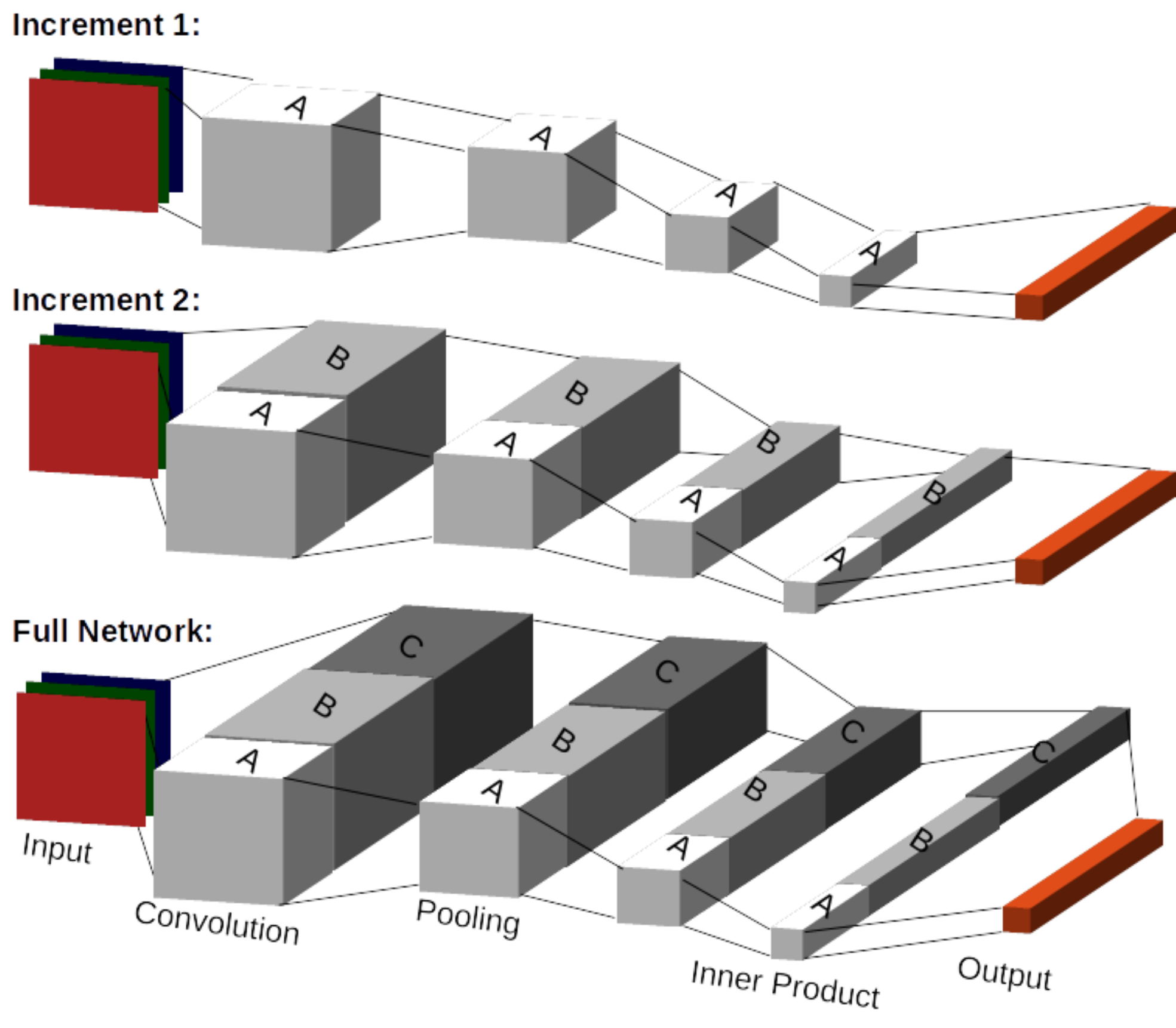}
    \vspace{-0.1in}
    \caption{Illustration of Incremental Training on a typical DNN.}
    \label{fig:net3d}
    \end{center}
    \vspace{-0.2in}
\end{figure}

Given a network, such as Figure \ref{fig:net}, we first form smaller or sub-networks from the original network by reducing the number of channels in each layer except for the output layer. For instance, the sub-networks labeled \textit{A} in Figure \ref{fig:net3d} is a smaller network created from the original network by keeping only channels labeled \textit{A} active while disabling those labeled \textit{B} and \textit{C}. During runtime, when only sub-network \textit{A} is used, all synaptic connections between channels in \textit{A} and those in \textit{B} and \textit{C} are cut, resulting in less computations in the forward pass, which translates directly to energy savings. However, since smaller networks are less accurate, it may be beneficial to have multiple sub-networks of different sizes such as \textit{A}, (\textit{A} $\cup$ \textit{B}) and the full network (\textit{A} $\cup$ \textit{B} $\cup$ \textit{C}) enabling the deployment of different network sizes as required. We keep the ratio of channels used in each sub-network compared to the full size network constant across all layers (excluding the last layer) to ensure feature representations are not lost between layers. In addition, this simplifies the search space for such sub-networks.

In order to allow sub-networks to be deployed independently while minimizing the weight storage requirement to that of a single network, we co-design the training algorithm (Algorithm \ref{alg:incr}) to train the network. We call this algorithm incremental training since the training process is done in increments. The input to Algorithm \ref{alg:incr} are the number (Num\_Incr) and the network architectures of the increments (Incr\_Arch), which contains the number of additional channels in each layer at different increments. Initially, we train the first increment, which is the smallest network (line 1 and 2 in the algorithm). Then, at each new increment, we expand the network by adding in more channels (line 4). We then train the resulting network while keeping all the weights in the previous training fixed (line 7). When a new channel is added, it is connected to all channels in the layers ahead and behind, so all these new synaptic connections are also trained. This process is repeated until the network size is equal to the that of the original network. Every time the network is trained with new channels, we keep a copy of the previous weights of the final output layer since these weights represent a unique output classifier. We refer to the number of trainings as the number of \textit{increments} in training process.

By performing incremental training, we provide the flexibility of using multiple networks of different sizes while storing and utilizing only one set of weights. As shown in Figure \ref{fig:net3d}, we can either deploy just the fraction \textit{A} or (\textit{A} $\cup$ \textit{B}) or the full network (\textit{A} $\cup$ \textit{B} $\cup$ \textit{C}). This deployment scheme allows for a trade-off between delay/energy and accuracy at runtime. We show next a systematic method to determine the optimal number and sizes of retraining increments for a given network.

\subsection{Network size versus inference accuracy} \label{ssec:numstep}
In order to effectively transform a given network so that it is runtime configurable, we first estimate the upper bounds of inference accuracy for various sizes of the network. For instance, we consider the accuracy achievable by the smaller network whose channels are labeled \textit{A} in Figure \ref{fig:net3d}. Then we do the same on the network whose channels are the union of \textit{A} and \textit{B}, and so on. This analysis is performed by the same method as incremental training except that in each training, we allow all weights to change. In addition, for each training, we use the same hyper parameters such as weight decay and momentum as given with the original network. We call these trained networks the Golden Models.

By gradually increasing/decreasing the size of the network, we have an estimate of how the number of channels affects the network accuracy. In section \ref{ssec:DCA}, we use this information to estimate the optimal number of retraining increments, which represents the number of networks with different sizes that could be independently deployed. For instance, when the number of retraining increments is 2, we can only deploy either a specific fraction of the network or the full network. Whereas when the number of increments is 3 as in Figure \ref{fig:net3d}, we can deploy either just \textit{A}, \textit{A} and \textit{B} combined, or the full network. The optimal number and sizes of increments would result in the lowest average size of network deployed per input using our algorithm outlined in Section \ref{ssec:DCA}.

\setlength{\textfloatsep}{4pt}
\begin{algorithm}[t!]
\label{alg:incr}
   \SetKwInOut{Input}{Input}
    \SetKwInOut{Output}{Output}
    \Input{Num\_Incr, Incr\_Arch}
    \Output{Trained Network}
    net = initialize(Incr\_Arch[1])\\
    \tcp{Train all weights in net}
    net = train(net, KEEP\_FIXED(NULL))\\
    \For{i = 2 to Num\_Incr}
      {
      \tcp{Add more channels and initialize their}
      \tcp{weights}
      tmp\_net = \{net $\cup$ initialize(Incr\_Arch[i])\}\\
      \tcp{Keep all weights corresponding to net}
      \tcp{fixed}
      tmp\_net = train(tmp\_net, KEEP\_FIXED(net))\\
      net = tmp\_net
      }
    \Return net
    \caption{Incremental Training}
\end{algorithm}

\subsection{Weight Initialization} \label{ssec:challenge}
Most DNNs are trained using the backpropagation algorithm with stochastic gradient descent or one of its variants. During training, the inference error from the output layer is propagated backward in the form of partial gradients, and the synaptic weights in each layer are updated concurrently. This training scheme presents a challenge for our method because in incremental training, we optimize the network according to only a subset of the weights at each training increment, and these weights are fixed in the next increment. Fortunately, the non-convex nature of the cost function in DNNs allows for many local optima, which could be very close to the local optimum found using the original training scheme. At each step, incremental training searches for these local optima by adapting features learned by new channels to features which are already captured by the fixed \textit{a priori} channels.

With a more restricted search space compared to the original training procedure, our method could lead to some accuracy drop in the full network. However, we found that the number of retraining increments and sizes of channel increments directly correlate with the final accuracy difference from the traditional training scheme. In particular, with few increments and large increment size, there is no accuracy drop. Depending on deployment scenario, designers can trade off the number of increments and accuracy of the network. We propose next, an initialization technique that helps lead to smaller drops in accuracy.

Good initialization of synaptic weights plays a key role in the success of training DNNs~\cite{initialization}. Given a network architecture, we first train a separate model for each increment, where all weights and biases are allowed to change, using the original initialization and training procedure. We call these the golden models. In Figure \ref{fig:net3d}, the golden models would be (\textit{A} $\cup$ \textit{B}) for the second increment, and (\textit{A} $\cup$ \textit{B} $\cup$ \textit{C}) for the third and last increment. Then, in incremental training, we initialize each new increment with its corresponding golden model and substitute in the fixed weights from the previous increment.

In Section~\ref{sec:Results}, we will evaluate the proposed initialization techniques and show that it leads to a significant boost in accuracy compared to regular incremental training.

\section{Runtime Methodology}
\label{sec:Runtime}
The increase in popularity of mobile platforms imposes strict regulations on delay and energy requirements. While DNNs are leading the state-of-the-art in accuracy performance, they are especially hard to be deployed in mobile settings due to their energy and high throughput requirements. We aim to provide a method, which can help designers achieve the accuracy goal but with smaller energy, delay and storage overhead. This system can be deployed in either of the two following schemes to achieve energy savings.

\begin{algorithm}[b!]
\label{controller}
   \SetKwInOut{Input}{Input}
    \SetKwInOut{Output}{Output}

    \Input{Constraints: $Energy$ and  $Delay$}
    \Input{System Performance: $sysEnergy$ and  $sysDelay$}
    \Output{$capacity$}
    \eIf{Energy or Delay changes}
      {
      \eIf{Energy or Delay decreased}
      	{
        	\If{$sysEnergy>Energy$}
            {
        		Decrease $capacity$;
            }
      	}
        {
        	$capacity=Max  Capacity$;
        }
      }
      {
    \If{$sysEnergy>Energy$}
        {
        	Decrease $capacity$;
        }
      }
    \caption{Feedback Controller for Dynamic Network Configuration}
\end{algorithm}

\subsection{Runtime, Energy and Delay Constraints} \label{ssec:RTE}
In this scheme, the system is given energy and delay constraints in real-time, so the system must adjust the network size to meet the constraints. One possible scenario is that real-time constraints force a DNN to provide an answer within a smaller window of time as in the case of DNN accelerators deployed in autonomous vehicles, where a sudden, unexpected situation could force an on-chip DNN to make a decision within a tighter window of time. Another possible scenario is in mobile devices, where low-power modes can demand a DNN to reduce its nominal power consumption during run time.

\begin{figure}
	\includegraphics[scale=0.50]{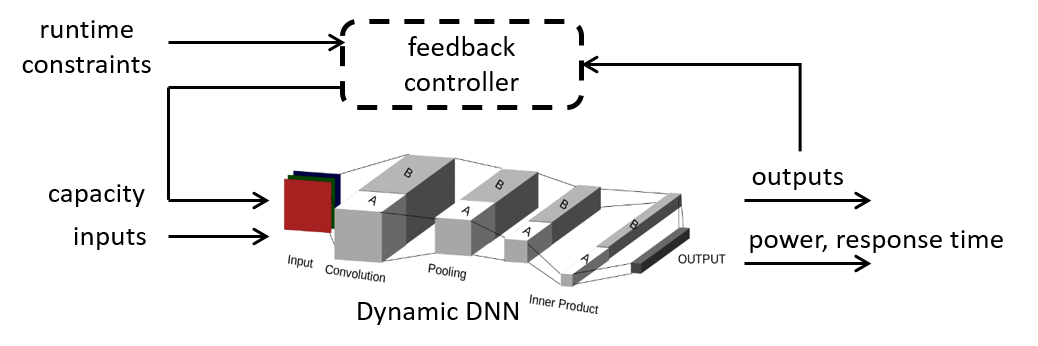}
    \vspace{-0.3in}
	\caption{Dynamic adjustment of DNN capacity using feedback controllers as implemented in the proposed constrained design approach. For real time constraints, the controller monitors the response time and power consumption of the DNN and adjusts its capacity based on the measurements and the target runtime constrains.}
	\label{fig:dynamicDNN1} 
    \vspace{0.15in}
\end{figure} 

We develop Algorithm \ref{controller} for our feedback controller given in Figure \ref{fig:dynamicDNN1} to regulate the number of channels or network capacity allowed at any given point, This algorithm first checks the energy and delay constraints and current system performance. If the energy and/or delay budget is not met, the network capacity is adjusted accordingly. At any point, the controller tries to adjust the capacity such that the system performance is close to, but does not exceed, the constraints. This allows for the highest possible inference accuracy while not violating the constraints. However, when the constraints loosen (i.e., allowing higher energy or delay), in order to avoid implementing expensive controller circuitry with memory, we allow the controller to jump to the biggest network and then settle to the correct capacity based on the current constraints.

It is desirable, in this scheme, to maximize the number of retraining increments in the incremental training since this would allow more flexibility at runtime. However, large number of increments could lead to larger accuracy drop in the full network as discussed in Section \ref{ssec:challenge}. This results in a trade off, which would vary between applications. In this work, we use 4 retraining increments to demonstrate the resilience of incremental training.

\subsection{Opportunistic Energy Saving Scheme}

\label{ssec:DCA}
\begin{figure}
	\includegraphics[scale=0.50]{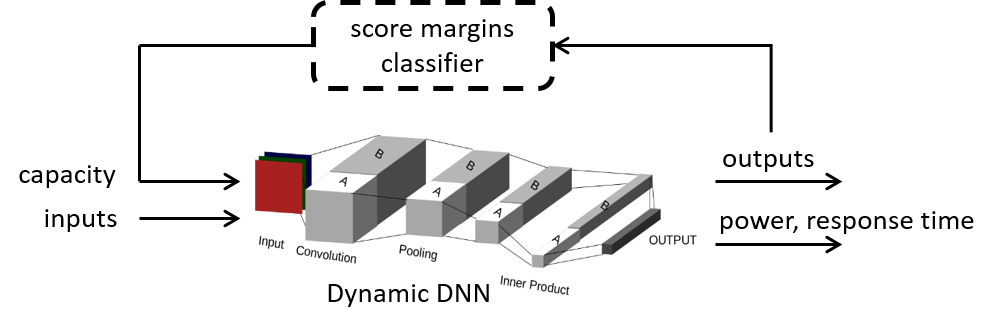}
    \vspace{-0.3in}
	\caption{Dynamic adjustment of DNN capacity using score margin classifiers as implemented in the proposed opportunistic approach. The score margin unit scales down the DNN to save energy as long as accuracy is not compromised.}
	\label{fig:dynamicDNN2} 
\end{figure}

In this scheme, our goal is to maximize energy saving while minimizing accuracy loss. This is equivalent to minimizing the average number of computations needed per input, which is achieved by deploying the smallest fraction of the network. However, since a smaller network is less accurate in general, there needs to be a recovery mechanism when the inference of the smaller network is wrong. As shown in Figure \ref{fig:dynamicDNN2}, our technique consists of a runtime configurable DNN of choice and a score margin classifier.

Score margin is defined as the absolute difference between the two largest neuron outputs (scores) in the final layer of a DNN.Leveraging the observations from Park \textit{et al.}~\cite{park} that there is a strong correlation between the top two score margins and the prediction accuracy, we use this margin information in our recovery mechanism.

\setlength{\textfloatsep}{10pt}
\begin{algorithm}[h]
\label{alg:opportunistic}
   \SetKwInOut{Input}{Input}
    \SetKwInOut{Output}{Output}
    \Input{TrainedNet, ImageIn, Threshold}
    \Output{Class}
    ForwardPass() $\triangleq$ performs a forward pass evaluation on the input\\
    Net = TrainedNet[increment=1] \\
    (Class, ScoreMargin) = ForwardPass(Net,ImageIn)\\
      \While{ScoreMargin < Threshold[increment]}{
      increment = increment + 1\\
      Net = TrainedNet[increment]\\
      (Class, ScoreMargin) = ForwardPass(Net,ImageIn)\\
      \If{TrainedNet[increment+1]=NULL}{ break }
      
      }
     \Return Class
    \caption{Runtime Opportunistic Energy Saving Scheme}
\end{algorithm}

In this approach, when the top two score margins fall below a certain threshold, we deploy a bigger fraction of the network. Algorithm \ref{alg:opportunistic} illustrates this process. This threshold can be set statically or adjusted dynamically. Compared to Park \textit{et al.}, the memory requirement for our method is significantly less since we only need to store small additional weights for the final output layer for various-sized networks. The difference becomes even bigger when the number of retraining increments increases, as will be shown in Section~\ref{ssec:experimental}. In addition, we provide a systematic search approach for such networks.

\begin{figure}[b]
	\vspace{-0.025in}
	\includegraphics[scale=0.6,trim=10 0 10 5,clip]{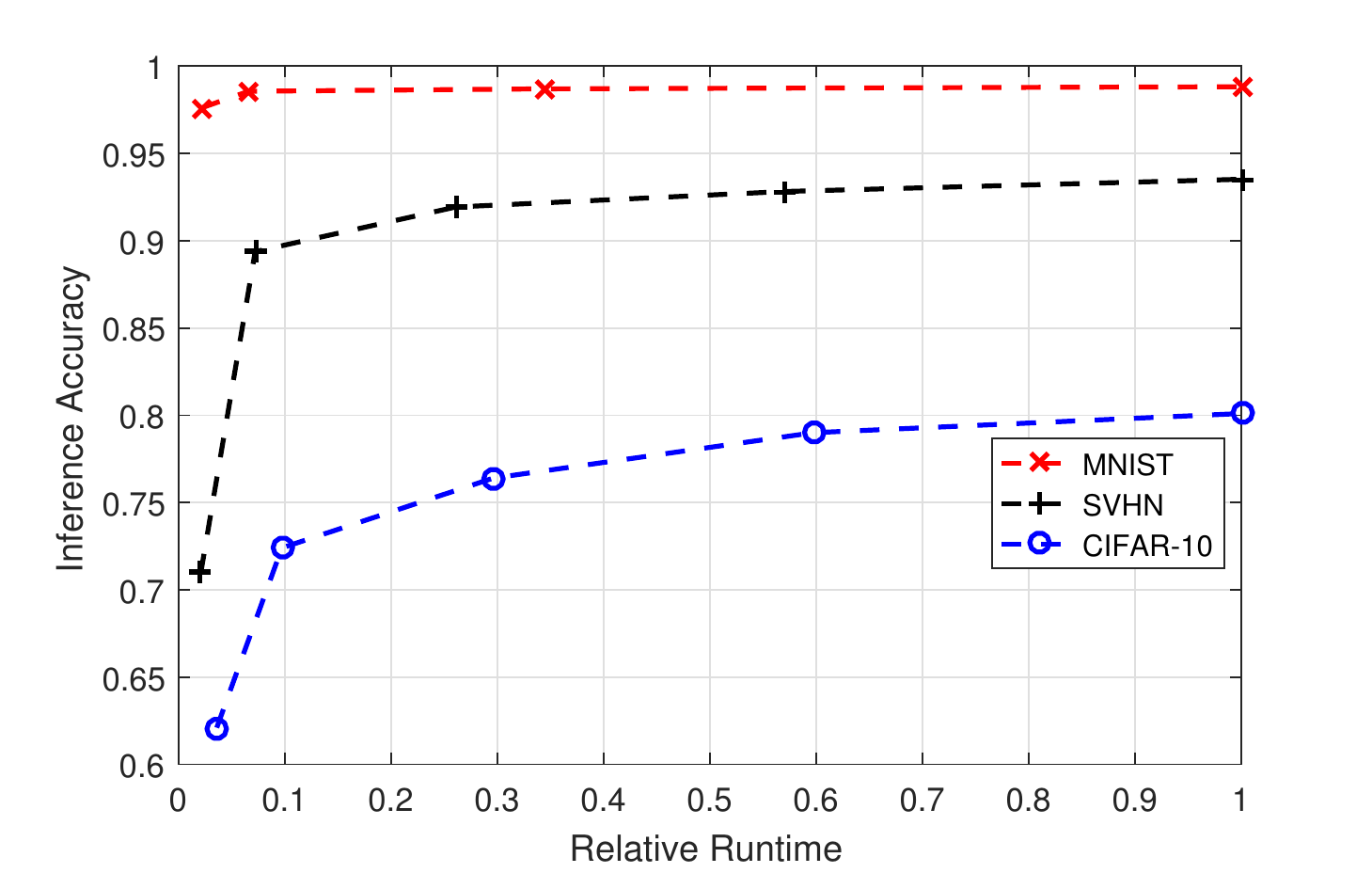}

    \vspace{-0.1in}
	\caption{Inference accuracy of golden model in validation set versus relative network runtime in forward pass.}
	\label{fig:goldenacc}
    \vspace{-0.1in}
\end{figure}

The optimal number of retraining increments is not necessarily the largest one in this scheme. As discussed below, it depends on the accuracy of the full network, the initial increment and the accuracy increase of each increment. Let $E$ be the expected fraction of network deployed per input. We can compute $E$ as follows:
\begin{align} \label{eq:fraction}
	E=\sum^{N}_{i=1}\left[\sum^{i}_{j=1}f_j\right]\cdot&[P(SM_i > \theta_i | f_i)\\\nonumber
    				&-P(SM_{i-1} > \theta_{i-1} | f_{i-1})],
\end{align}
where $f_i$ represents the fraction of the network in increment $i$, $N$ is the number of increments until the full network is deployed, and $P(SM_i > \theta_i | f_i)$ denotes the probability that the score margin ($SM$) is greater than the threshold $\theta_i$ in increment $i$ given that the network size is $f_i$, so the inference result is final. At increment $N$, the full network is deployed, so $f_N=1$, $\theta_N=0$ or undefined, and $P(SM_N > \theta_N | 1)\triangleq 1$. For $i<N$, $P(SM_i > \theta_i | f_i)$ can be approximated as follows:
\begin{align}
	P(SM_i &> \theta_i | f_i) = P(f_i\text{ correct})\cdot P(SM_i>\theta_i|f_i \text{ correct})\nonumber\\
			+ &(1-P(f_i\text{ correct}))\cdot P(SM_i > \theta_i | f_i \text{ wrong})),
\end{align}
where $P(f_i\text{ correct})\in [0,1]$ is the accuracy of $f_i$. Figure \ref{fig:goldenacc} shows the inference accuracy of the golden models for the three networks we considered for this paper. We obtain $P(f_i\text{ correct})$ by curve fitting the accuracy versus network size using this figure.

The expected accuracy of the network deployed using our incremental method ($net_{acc}$) is computed by:
\begin{equation} \label{eq:netacc}
net_{acc}=
\begin{cases} 
		T_1 &\mbox{if }N>1\\
		P(f_N \text{ correct}) &N=1, 
\end{cases}
\end{equation}
where
\begin{equation*}
T_{i}= 
\begin{cases}
		P(SM_i<\theta_i|f_i\text{ correct}) &\mbox{if }i<N\\
			\hspace{0.3in}+ P(SM_i<\theta_i)\cdot T_{i+1} \\
		P(f_N \text{ correct}) &\mbox{otherwise}.
\end{cases}
\end{equation*}
Our goal is to choose $\bf{F}$=$(f_1,f_2,\ldots,f_N)$, $\bf{\Theta}$ = $(\theta_1,\ldots,\theta_N)$, and $N$ so as to minimize $E$ while maximizing $net_{acc}$. This can be stated as:
\begin{equation} \label{eq:optimize}
	argmin_{N,\bf{F},\bf{\Theta}}\left[E, 1-net_{acc}\right].
\end{equation}

\begin{figure}[ht]
	\includegraphics[scale=0.55,trim=20 8 22 10,clip]{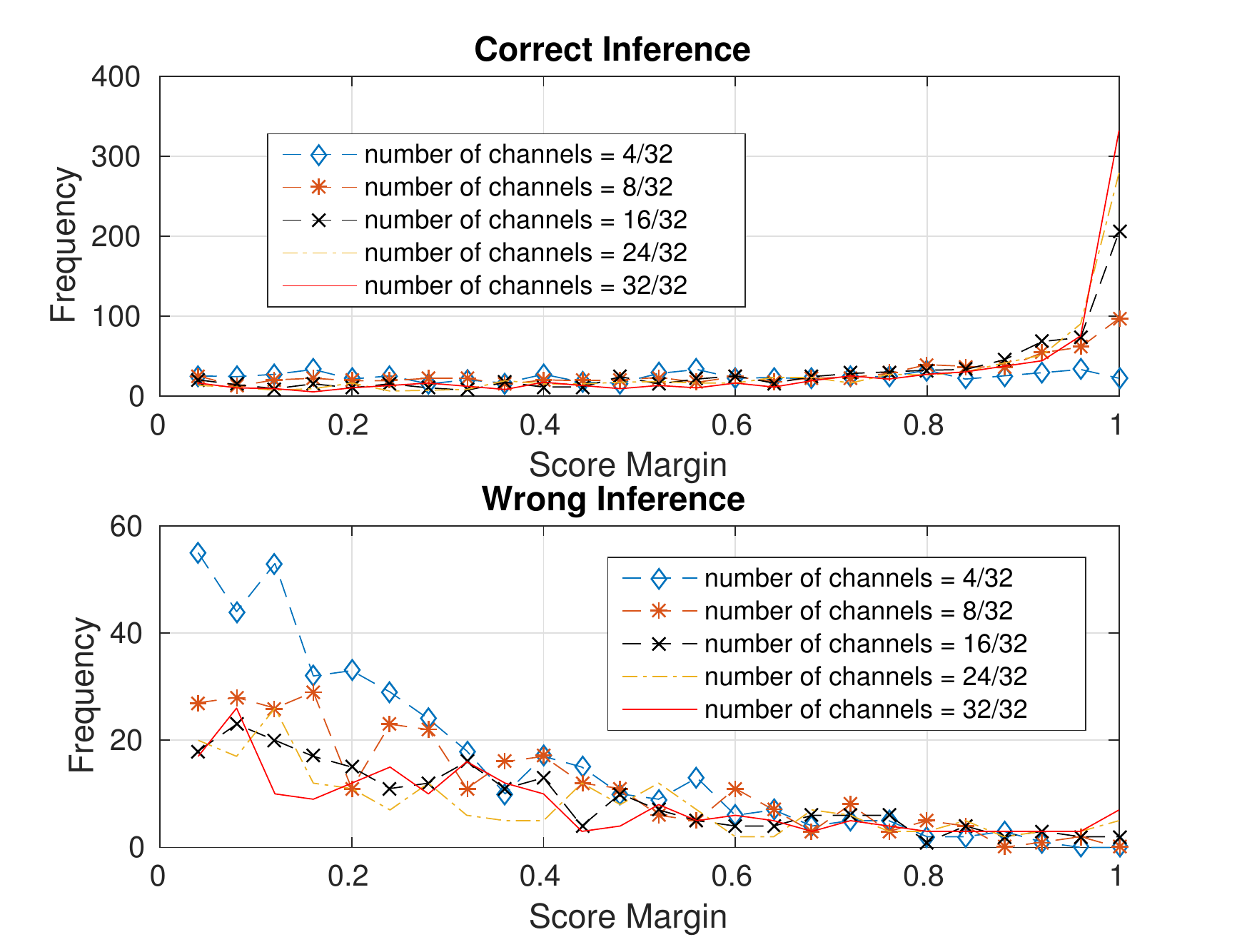}
	\caption{Histogram for top two class scores margin for correct inference (top) and wrong inference (bottom) for CIFAR-10 dataset. The number of channels (x/y) shows the ratio of number of channels in the first layer of the network in use (x) and that of the full network (y). This ratio is identical for all layers except the final layer.}
	\label{fig:cifarscoremargin}
\end{figure}

We observe that, for any $\theta_i\in[0,1]$, $P(SM_i<\theta_i|f_i$ correct) increases as $f_i$ decreases, as shown in the top plot of Figure~\ref{fig:cifarscoremargin}.  This is beneficial for $net_{acc}$; however, the energy savings is not optimal in this case because larger networks could be deployed even though the inference of the smaller network is correct. To analyze score margin behavior for different network sizes, we perform a forward pass on the validation set using the golden models in Figure \ref{fig:goldenacc} and report the score margin in Figure~\ref{fig:cifarscoremargin}. Given limited space, we only report the result for CIFAR-10. Based on these results, training with $f_i$ too small can in fact hurt $E$ for $\forall\theta_i \in [0,1]$; e.g. when the number of active channels is 4, the score margin for the correct inference case has no clear trend. Thus, we set a minimum value for $f_i$ for each benchmark network.

Figure \ref{fig:cifarscoremargin} also shows that $\theta_i$ correlates with $f_i$, so for each $f_i$, we can use the static method used by Park \textit{et al.}~\cite{park} to find $\theta_i$ such that Eqn.~(\ref{eq:optimize}) is optimized. Thus, we first compute optimal $N$ and $f_i$ by optimizing Eqn.~(\ref{eq:fraction}). Keeping the fraction of active channels uniform across layers in the network allows a relatively small search space for $N$ and $f_i$. This uniform fraction is also necessary for preserving the information between layers. Thus, we can optimize Eqn.~(\ref{eq:fraction}) by simply sweeping through all possible values of $N$ and $f_i$ assuming that $P(SM_i<\theta_i|f_i\text{ correct})=P(SM_i\ge\theta_i|f_i\text{ wrong})=0$.

\section{Experiments}
\label{sec:Results}

\subsection{Experimental Setup}

We evaluate our proposed techniques on two different platforms: (1) a custom hardware accelerator and (2) a low-power embedded GPU.  Details of our evaluation platforms follow.\\

\begin{figure}[b]
\vspace{-0.1in}
   \begin{center}
    \includegraphics[scale=0.34,trim=5 7 5 7,clip]{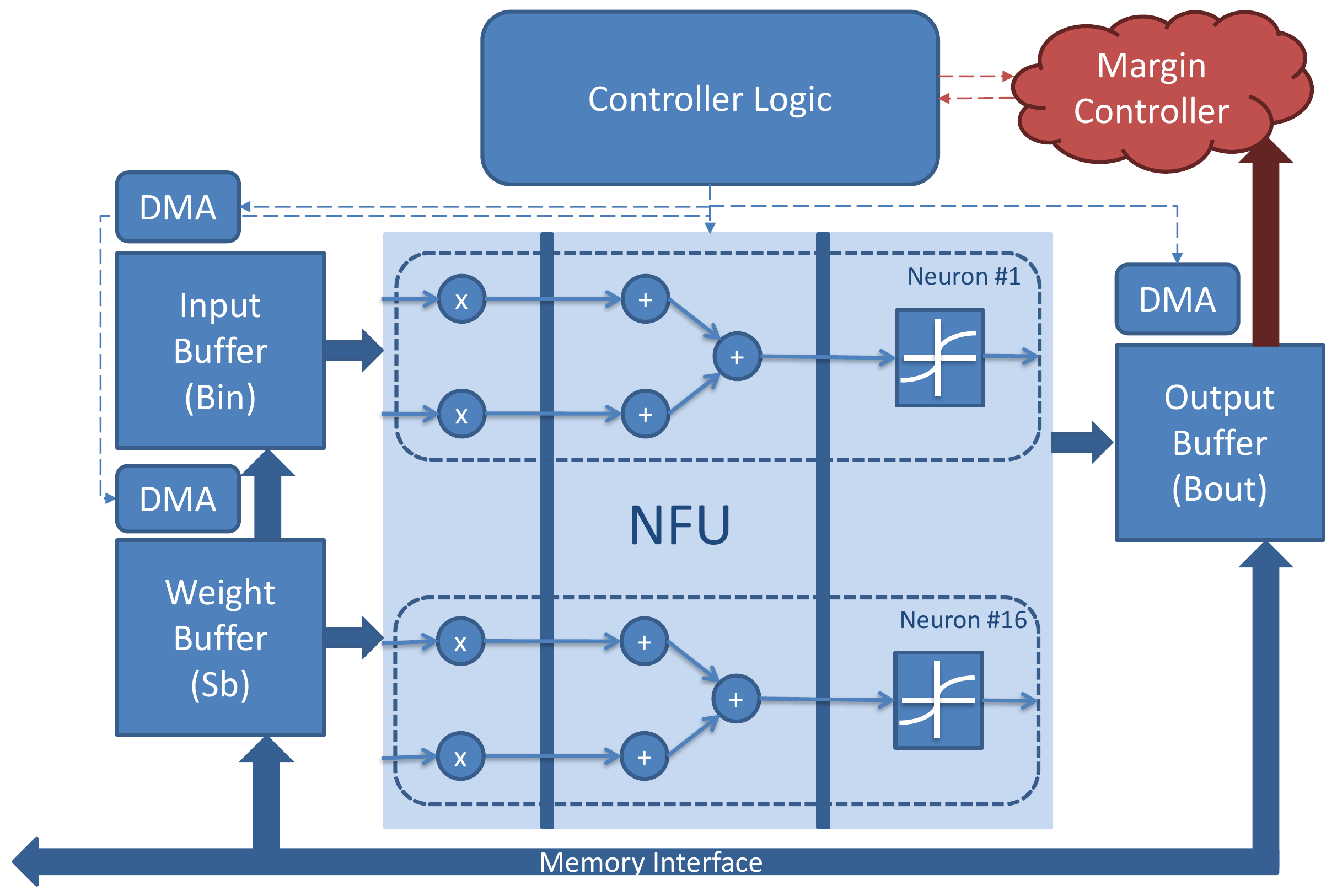}
    \caption{The custom HW implemented in our work.}
    \label{fig:hw}
    \end{center}
    \vspace{-0.15in}
\end{figure}

\noindent{\bf 1. Custom Hardware Accelerator.} For our custom accelerator experiments, we adopt a tile-based design similar to DianNao~\cite{chen}.  We use a 65 nm technology node and Synopsys Design Compiler to synthesize our design. We implement 16 neuron processing units, with 16 synapses for each neuron where the inputs, weights, and outputs are stored in separate SRAM buffers for high throughput. We use 32-bit fixed point arithmetic and the calculation of the output for each neuron is divided into three phases where the phases are multiplication, additions, and the non-linearity functions. Figure~\ref{fig:hw} illustrates the organization of our accelerator design.  We also design and implement our own custom hardware controller which enables the accelerator to calculate and utilize the score margin for each image and decide whether to move to the next image or rerun the same image on a bigger network. In Figure~\ref{fig:hw} we highlight this controller logic as margin controller, which differentiates our implementation from DianNao. Our controller adds an insignificant area and power overhead of approximately 0.15\%, and delay overhead of 12ns each time it is activated. In our implementation, the total capacity of our three buffers is 90KB. We model an off-chip DDR3 DRAM memory with 4GB of capacity for storing the weights. To evaluate the design metrics and external memory, we use CACTI~\cite{CACTI} models. 

To isolate the inaccuracies introduced using the proposed methods and to eliminate the quantization inaccuracies, we use a bit-width that ensures there is no degradation in accuracy compared to floating point as implemented in software. We observe, empirically, that in our applications, fixed point representation using 32-bits delivers no drop in accuracy compared to floating point while offering some benefits in design parameters. Therefore, for Section~\ref{ssec:experimental}, we use a 32-bit fixed point implementation to report the performance. We provide in Table~\ref{table:hw_comp} the design characteristics of different bit-widths as well as a single precision floating point implementation. Here, $\langle n,f\rangle$ shows the bit-width and the fraction part width respectively. As expected, as we increase the bit-width, the area and power increase. Also, the floating point implementation has the highest delay and  power overhead. Table~\ref{table:hw} provides in detail the hardware characteristics of the 32-bit fixed point hardware implementation. The resulting hardware is capable of performing 496 fixed point operations every clock cycle resulting in 124 GOP per second or 102.48 GOP per Watt.\\

\begin{table}
  \small
  \centering
  \begin{tabular}{c|c|c|c}
    \hline
    Design &  Area & Power &  Delay \\
       & ($mm^2$) & ($mW$) & ($ns$)\\
    \hline \hline
     Floating Point & \multicolumn{1}{r}{16.74} & \multicolumn{1}{r}{1379.6} & \multicolumn{1}{r}{3.99} \\
     Fixed Point <32,16> & \multicolumn{1}{r}{14.13} & \multicolumn{1}{r}{1213.4} & \multicolumn{1}{r}{3.99}  \\
     Fixed Point <16,10> & \multicolumn{1}{r}{6.88} & \multicolumn{1}{r}{574.8} & \multicolumn{1}{r}{3.94}  \\
    \hline
  \end{tabular}
  \vspace*{-1mm}
  \caption{The hardware implementation characteristics of different arithmetic and bit-widths.}
  \label{table:hw_comp}
\end{table}

\begin{table}[t]
  \scriptsize
  \centering
  \begin{tabular}{c|r|r|r|r}
    \hline
    Component &  \multicolumn{1}{|c}{Area} & \multicolumn{1}{|c}{Area} & \multicolumn{1}{|c}{Power} & \multicolumn{1}{|c}{Power}  \\
      		  & \multicolumn{1}{|c}{($um^2$)} & \multicolumn{1}{|c}{(\%)} & \multicolumn{1}{|c}{($mW$)} & \multicolumn{1}{|c}{(\%)}\\
    \hline \hline
     Total & 14,133,270 &    & 1213.4  \\ \hline
     Combinational & 1,220,249  & 8.63 & 140.4 & 11.57  \\
     Buffer/Inverter & 76,716 & 0.54 & Neg. & 0  \\
     Registers & 14,133,270 & 0.65 & 28.0 & 2.31 \\
     Memory & 14,133,270 & 90.72 & 1,044.9 & 86.11  \\ \hline
     Sb (Weights Buffer) & 11,369,714 & 80.64 & 979.96 & 80.76  \\
     Bin (Input Buffer) & 712,294 & 5.04 & 61.25 & 5.05  \\
     Bout (Output Buffer) & 712,295 & 5.04 & 61.25 & 5.05  \\
     NFU (Functional Unit) & 1,275,780 & 9.03 & 163.85 & 13.50  \\
     Control Logic & 36,187 & 0.26 & 4.06 & 0.33  \\
    \hline
  \end{tabular}
  \caption{Breakdown of hardware implementation characteristics of different components.}
  \vspace{-0.1in}
  \label{table:hw}
\end{table}

\noindent{\bf 2. Embedded GPGPU:} For our GPU experiments, we use single precision floating point weights and inputs and evaluate our techniques using the Nvidia Jetson TX1 board. We choose an embedded GPU to demonstrate the critical improvement in runtime and energy of our method when DNNs are deployed in a resource constrained environment. The board features a quad-core 64-bit ARM A57 CPU and a 256-core Nvidia Maxwell GPU. We modify Caffe \cite{caffe} to take advantage of our dynamic network configuration method.\\

\noindent{\bf Benchmarks.} All of our experiments are performed using three well-known DNNs, for MNIST, CIFAR-10\footnote{We remove the local response normalization layers from the AlexCIFAR-10 network to simplify our hardware implementation. In agreement with recent literature, which questions the necessity of such a layer, we found that the accuracy drop is small (less than 1\%).} and SVHN datasets~\cite{mnist, cifar10, svhn}. Benchmark details are given in Table~\ref{table:networks}.  We do not perform pre-processing on any of the datasets other than normalization or mean subtraction. For MNIST and CIFAR-10, we randomly split out 10\% of each classification category from the original test set as our validation set. For the SVHN dataset, we prepare the validation and training sets using the same method as Sermanet \textit{et al.} \cite{convnet} except that we do not preprocess the images. We chose the three networks to reflect varying final network accuracies and to illustrate the importance of multi-step flexibility through incremental training. Our networks are trained using Caffe~\cite{caffe}. Our analysis in Section \ref{sec:Methodology} is performed on the validation sets, and we report all results in this section from the test sets.

\begin{table}[t!]
  \normalsize
  \centering
  \begin{tabular}{c|c|c}
    \hline
    LeNet \cite{lenet} & ConvNets \cite{convnet} &  ALEXnet \cite{cifar10} \\
    \hline
    28$\times$28 & 32$\times$32$\times$3 & 32$\times$32$\times$3 \\
    conv 5$\times$5$\times$20 & conv 5$\times$5$\times$16 & conv 5$\times$5$\times$32 \\
    maxpool 2$\times$2 & avgpool 2$\times$2 & maxpool 3$\times$3 \\
    conv 5$\times$5$\times$50 & conv 7$\times$7$\times$512 & conv 5$\times$5$\times$32 \\
    maxpool 2$\times$2 & conv 5$\times$5$\times$20 & avgpool 3$\times$3 \\
    innerproduct 500 & avgpool 2$\times$2 & conv 5$\times$5$\times$64 \\
    innerproduct 10 & innerproduct 20 & avgpool 3$\times$3 \\
                    & innerproduct 10 & innerproduct 10 \\
    \hline
  \end{tabular}
  \caption{Benchmark Networks Architecture Descriptions.}
  \label{table:networks}
\end{table}

\subsection{Experimental Results} \label{ssec:experimental}
In this section, we first demonstrate the strength of incremental training by comparing its accuracy performance to channels shutdown, where we shutdown a number of channels from each layer of the full network for each input. The number of channels left active in each layer is equal to that of the incremental training counterpart to ensure that the number of computations needed are the same for the two networks. In addition, we show the effect of the initialization procedure proposed in Section \ref{ssec:challenge}.

\begin{figure}[t!]
   \begin{center}
    \includegraphics[scale=0.56,trim=28 15 28 25,clip]{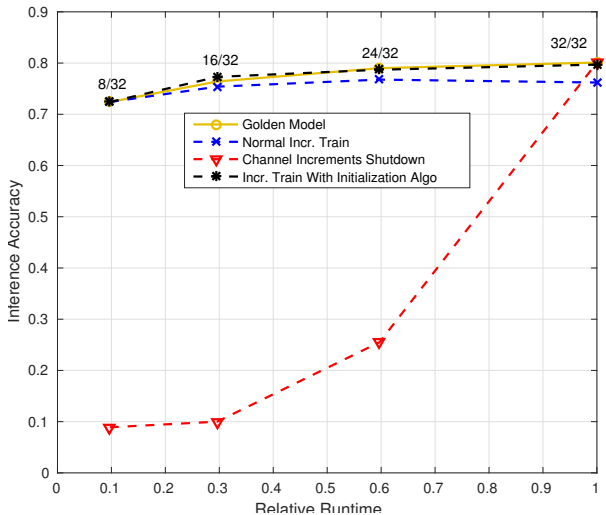}
    \vspace{-0.15in}
    \caption{Comparison of inference accuracy on CIFAR-10 validation set for golden model, incremental training, channel increments shutdown and incremental training with initialization from Section \ref{ssec:challenge}. Relative runtime is the ratio of the forward-pass runtime to that of the full network. The two numbers displayed at each datapoint (x/y) shows the number of channels as explained in Figure \ref{fig:cifarscoremargin}.}
    \label{fig:effects}
    \end{center}
    \vspace{-0.15in}
\end{figure}

To demonstrate the strength of incremental training and the initialization procedure, we train ALEXnet DNN using these two methods. We then compare their accuracy performance to that of the golden model as shown in Figure \ref{fig:effects}. However, deployment of the golden model is unrealistic since it would require extra storage for each network of different sizes. For a fair comparison, we also perform channel increments shutdown experiment, where we shutdown channels in the full network of the golden model. The number of training increments is set to four for all experiments. After the first increment (8/32 in the figure), a large fraction of the weights is kept fixed at each new training increment, yet the accuracy continues to increase for incremental training. On the other hand, channel increments shutdown result in disastrous accuracy drop. This highlights the importance of incremental training and its resilience despite the small fraction  of trainable weights. In normal incremental training mode, however, the accuracy drops when going from the third to the last increment (from 24/32 to 32/32 in the figure). It is observed that this drop is due to the large magnitudes of the weights in the the third increment, which are kept fixed, interfering with the learning of new weights in the fourth increment, especially since these new weights are normally initialized to very small values. This accuracy drop reduces significantly when we apply the initialization procedure proposed in Section \ref{ssec:challenge}, which initializes the new weights to comparable magnitudes to the fixed ones. In addition, initialization from the golden models helps ensure a good starting point.

As demonstrated in Figure~\ref{fig:effects}, weight initialization results in considerable accuracy boost for incremental training and therefore, we perform our training using this method for the rest of our experiments. Next, we report the energy savings and accuracy results for the two scenarios as described in Section \ref{sec:Runtime}.

\begin{figure*}[ht]
   \begin{center}
    \includegraphics[scale=0.6, trim=85 2 85 3,clip]{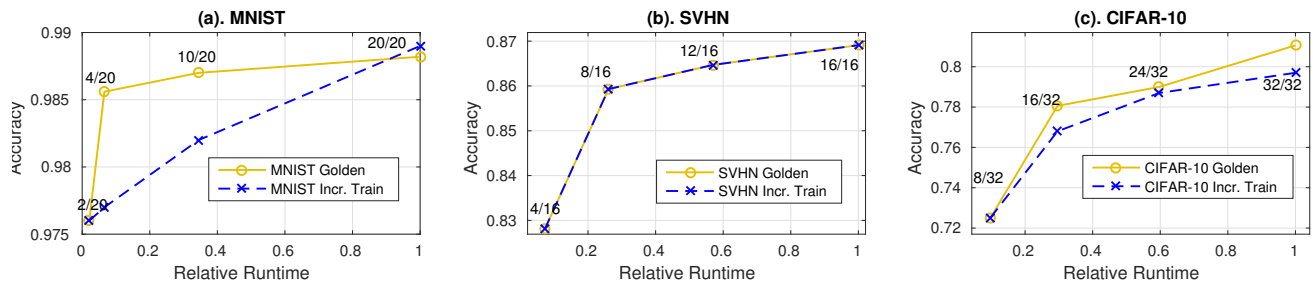}
    \vspace{-0.05in}
    \caption{Test set inference accuracy versus network relative runtime for (a) MNIST, (b) SVHN, and (c) CIFAR-10. The two numbers (x/y) at each data point have the same representation as Figure \ref{fig:cifarscoremargin}.}
    \label{fig:incremental}
    \end{center}
    \vspace{-0.2in}
\end{figure*}

\noindent\textbf{1. Runtime Energy, Delay Constraints:} For these experiments, first we incrementally train each network to support 4 different increments. As discussed in previous sections, increasing the number of increments without provisions  results in significant accuracy losses. Therefore, selecting the number of increments is a trade-off between the reduction in accuracy and the flexibility to perform within close vicinity of the constraints. We choose 4 increments due to the fact that a larger number of increments results in further accuracy drop even when the full network is deployed. The achieved accuracies for each network and for each network size is shown in Figure~\ref{fig:incremental}. Also in Figure~\ref{fig:incremental} we compare the accuracies of our incrementally trained networks to their respective golden models, which has a weight set for each increment to avoid accuracy loss, as described in training initialization in Section \ref{sec:Methodology}. 

Figure \ref{fig:incremental} demonstrates that we are able to transform our benchmark networks into runtime configurable with 4 increments while sacrificing a maximum of 1.4\% of full network accuracy. It is critical to note that this 1.4\% reduction in accuracy is due to high number of increments. It is up to designers to decide on the trade off. With small number of increments, the reduction is negligible, as shown in the Opportunistic Energy Saving scheme. We also evaluate the energy and delay characteristics of our proposed methods using two domains commonly used within the embedded system design framework. In Table \ref{table:energy}, we give the energy costs and response time when deploying each increment for each of our three networks using our custom accelerator, while Table~\ref{table:energy2} summarizes the results when the networks are implemented on the TX1 GPU board. Table \ref{table:energy} reports good saving when going from the first increment to the fourth. However, there are some inconsistencies in Table \ref{table:energy2}, where some smaller networks have larger execution time/energy than the larger ones. Our profiling results show that Caffe maps the smaller networks to less optimized GPU kernels in the CuDNN library\footnote{https://developer.nvidia.com/cudnn}. In addition, some kernels appear to be the bottlenecks as they have similar runtime for smaller and larger networks. Thus, the execution time/energy difference among the various increments in Table \ref{table:energy2} is a modest estimate.

\begin{table}[t]
  \centering
  \scriptsize
  \begin{tabular}{c|r|r|r|r|r|r}
    \hline
     & \multicolumn{2}{c|}{LeNet} & \multicolumn{2}{c|}{ConvNets} & \multicolumn{2}{c}{ALEXnet } \\
    Incr. & T($us$) &  E($uJ$) & T($us$) & E($uJ$) & T($us$) & E($uJ$) \\
    \hline
    \hline
    1 & 4.61 & 5.59 & 86.49 & 103.78 & 57.80 & 70.12 \\
	2 & 14.43 & 17.48 & 313.69 & 376.41 & 177.04 & 214.80 \\
    3 & 75.54 & 91.61 & 686.51 & 823.76 & 357.74 & 434.03 \\
    4 & 283.04 & 343.39 & 1203.34 & 1443.90 & 599.85 & 727.81 \\
    \hline
  \end{tabular}
  \vspace{-0.1in}
  \caption{Mean energy cost (E) and processing time (T) per input image when different fractions of the each networks are deployed using our custom hardware accelerator.}
  \label{table:energy}
\end{table}

\begin{table}[t]
  \centering
  \scriptsize
  \begin{tabular}{c|r|r|r|r|r|r}
    \hline
     & \multicolumn{2}{c|}{LeNet} & \multicolumn{2}{c|}{ConvNets} & \multicolumn{2}{c}{ALEXnet } \\
    Incr. & T($us$) &  E($uJ$) & T($us$) & E($uJ$) & T($us$) & E($uJ$) \\
    \hline
    \hline
    1 & 1.64e03 & 1.40e04 & 1.97e03 & 1.45e04 & 2.52e03 & 2.09e04 \\
	2 & 1.81e03 & 1.58e04 & 2.45e03 & 2.41e04 & 2.46e03 & 2.01e04 \\
    3 & 2.23e03 & 1.88e04 & 2.25e03 & 2.00e04 & 2.77e03 & 2.16e04 \\
    4 & 3.14e03 & 2.94e04 & 3.22e03 & 3.05e04 & 3.28e03 & 3.13e04 \\
    \hline
  \end{tabular}
    \vspace{-0.1in}
  \caption{Mean energy cost (E) and processing time (T) per input image when different fractions of the each networks are deployed using Nvidia Jetson TX1 GPU board. Input images are fed into the network one at a time.}
  \label{table:energy2}
\end{table}

\begin{figure}[t!]
   \begin{center}
    \includegraphics[scale=0.55,trim=0 0 0 10,clip]{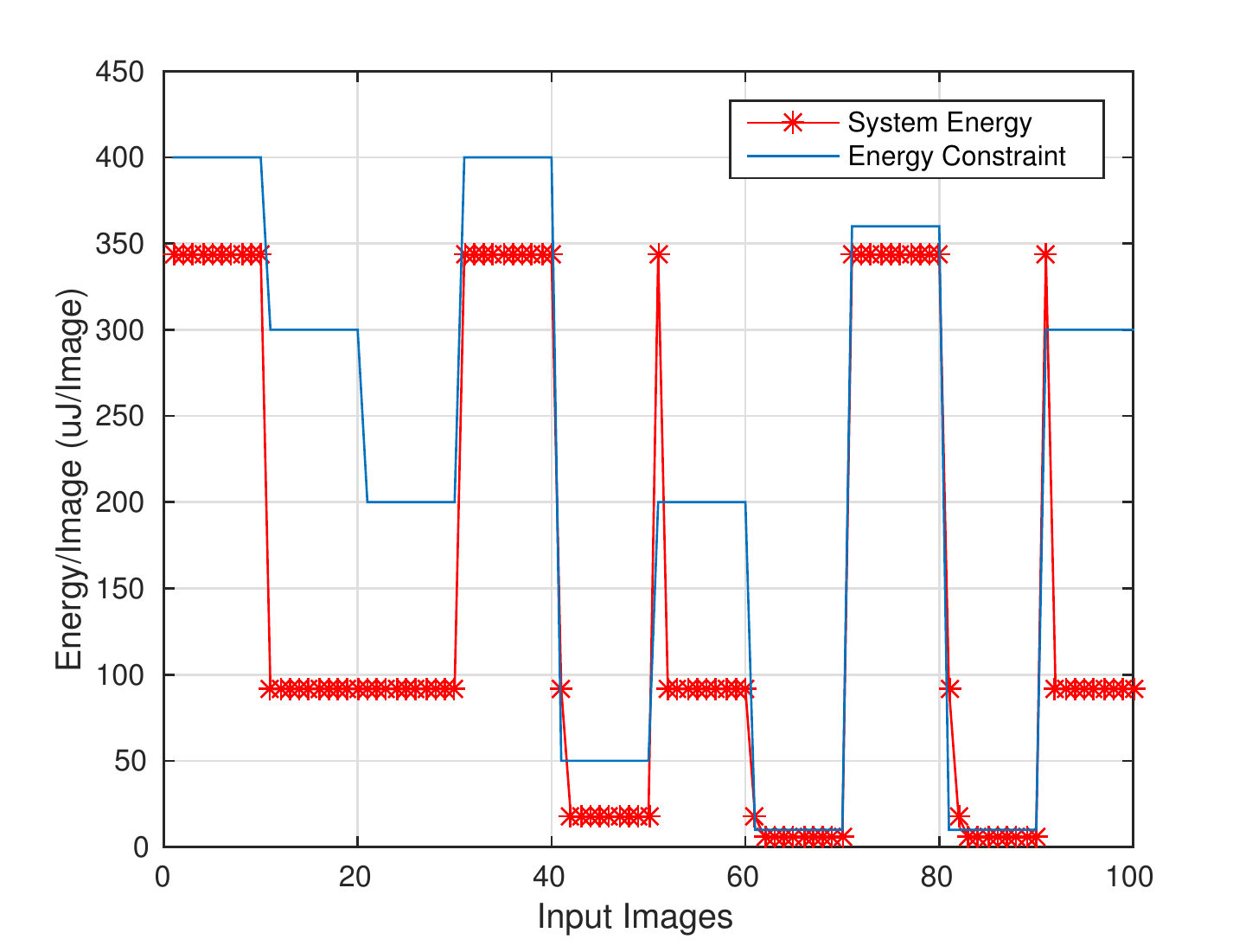}
    \vspace{-0.2in}
    \caption{Comparison of network energy adjusts with the imposed energy budget over time running MNIST tesebench.}
    \label{fig:controller}
    \vspace{-0.1in}
    \end{center}
\end{figure}

With this trained network, we next develop a set of runtime energy constraints to show the effectiveness of our runtime controller, as described in Algorithm \ref{controller}. Figure \ref{fig:controller} shows the controller in action, as implemented in our ASIC-based custom hardware, where we impose energy constraints during runtime, and the controller adjusts the network capacity to meet the constraints. We evaluate our controller on the MNIST network and the four levels of network energy per image are defined as summarized in Table~\ref{table:energy}. We see that with an incrementally trained network, the system is able to adapt to different energy requirements dynamically.

While in our work we focus on minimizing the memory requirements which would result in lower power consumption, in previous work, Park \textit{et al.}~propose to store two different networks and deploy them dynamically \cite{park}. This is also applicable to our case, where we simply store the golden models, each of which consists of four sets of weight each of different size. While this leads to a 1\% increase in the final network accuracy compared to incremental training, there is a heavy cost in storage requirement. Table \ref{table:memreq} provides a comparison between our method and Big/Little \cite{park}. As demonstrated in the table, saving different sets of weights rather than one, can result in up to a 96.43\% additional memory requirement in reference to the original network. In addition, during runtime these four different networks need to be in memory for fast dynamic switching, which would incur high memory power and could mask the energy saving from dynamic network configuration.\\

\begin{table}[t]
  \centering
  \begin{tabular}{c|r|r|r}
    \hline
     & LeNet & ConvNet &  ALEXnet \\
    \hline
    \hline
	Ours & 0.58\% & 0.10\% & 17.39\% \\
    Big/Little  \cite{park} & 30.71\% & 88.08\% & 96.43\% \\
    \hline
  \end{tabular}
    \vspace{-0.1in}
  \caption{Additional storage requirements normalized to the original network when the system is allowed to store multiple weights network.}
  \label{table:memreq}
\end{table}

\noindent\textbf{2. Opportunistic Energy Saving:} In this approach, for each DNN we first perform analysis on the optimal number and sizes of each increment, as discussed in Sections \ref{ssec:numstep} and~\ref{sec:Runtime}. Table \ref{table:increments} shows the computed optimal values using Eqn.~(\ref{eq:fraction}). Note that the fraction of active channels is uniform across all layers except the output layer, where the number of neurons is fixed. For instance, increment 1 of ALEXnet has 25\% of the number of channels in the full network.  Since the full network has 32 channels in its first layer, increment 1 will have 8 channels in its first layer. As shown in Table~\ref{table:increments},  when the network is highly accurate and resilient against large fractions of the channels disabled such as LeNet \cite{lenet}, the optimal number of increments is larger since larger savings can be achieved with a small probability of redoing the computation. Based on the results in Table \ref{table:increments}, we proceed to incrementally train the network. We then compute the optimal threshold $\theta$ for score margin for each network increment by maximizing the energy saving-accuracy product on the validation set. The threshold values are shown in parentheses.

\begin{table}[t]
  \centering
  \small
  \begin{tabular}{c|c|c|c}
    \hline
     & LeNet & ConvNets & ALEXnet  \\
    \hline
    \hline
    Num Incr. & 3 & 2 & 2 \\
	1st incr. & 0.25 $(0.90)$ & 0.25 (0.75) & 0.25 $(0.70)$ \\
    2nd incr. & 0.30 $(0.85)$ & 1 & 1 \\
    3rd incr. & 1 & \\
    \hline
  \end{tabular}
  \vspace{-0.1in}
  \caption{Optimal number of retraining increments for each network and fractions of active channels in each layer for each increment (score margin threshold in parentheses).}
  \label{table:increments}
\end{table}

\begin{table}[t]
  \centering
  \scriptsize
  \begin{tabular}{c|c|c|c|c|c|c}
    \hline
     & \multicolumn{2}{c}{LeNet} & \multicolumn{2}{c|}{ConvNets} & \multicolumn{2}{c}{ALEXnet } \\
    Incr. & Acc. &  E($uJ$) & Acc. & E($uJ$) & Acc. & E($uJ$)\\
    \hline
    \hline
    1 & 0.9760 & 5.59 & 0.828 & 103.78 & 0.724 & 70.12 \\
    & (0.9760) & & (0.828) &  & (0.724) & \\ \hline
	2 & 0.9866 & 17.48 & 0.8637 & 1443.90 & 0.8088 & 727.81\\
    & (0.9856) & & (0.8691) & & (0.8106) & \\ \hline
    3 & 0.9885 & 343.39\\
    & (0.9882) & \\ \hline
   \end{tabular}
   \vspace{-0.1in}
  \caption{Inference Accuracy (in parenthesis is the accuracy of the golden model for network with the same size as the increment) and energy cost for each increment in incremental training.}
  \label{table:incremental}
\end{table}

Table \ref{table:incremental} shows the inference accuracy and the energy of the different network increments. When the number of retraining increments is small, such as the case here, the accuracy difference between the incrementally and traditionally trained networks are almost negligible. Table \ref{table:incremental} shows that the maximum accuracy difference in the full network between the two training schemes is 0.52\%. Table \ref{table:opp} shows the energy saving and accuracy drops for each of our three benchmarks as evaluated on both of our platforms (i.e., the hardware accelerator and the TX1 GPU board). The GPU result of CIFAR-10 is omitted since the saving is minimal because, as discussed previously, the smaller network gets mapped to less optimized kernels. Compared to Big/Little \cite{park}, we are able to achieving the same or better saving with smaller memory requirements. We also provide a systematic method for achieving such saving.\\

\begin{table}[t]
  \normalsize
  \centering
  \begin{tabular}{c|c|c|c}
    \hline
     & LeNet & ConvNets & ALEXnet \\
    \hline
    \hline
    Accuracy Drop & 0.60\% & 0.96\% & 0.29\% \\
    Acc. Energy Savings & 95.53\% & 58.74\% & 32.61\% \\
	GPU Energy Savings & 48.00\% & 18.39\% & \\
    \hline
  \end{tabular}
  \vspace{-0.1in}
  \caption{Energy savings and accuracy drops for the dynamic configuration normalized to the golden result.}
  \label{table:opp}
\end{table}

\vspace{-0.1in}
\section{Conclusions}
\label{sec:conclusion}
The massive computational requirements of DNNs presents a challenge for its application on mobile platforms, where energy and delay budgets are restricted. However, with its state of the art accuracy, DNNs are becoming prevalent. In this work, we proposed a dynamic configuration approach for DNNs in conjunction with a co-designed incremental training methodology. Our approach achieves the targeted accuracy while allowing for runtime configurable energy and delay budget. It also enables the DNN to meet runtime constraints such as response time or power with graceful trade-off in accuracy. We show that our technique could be used to enable large energy saving with very small accuracy reduction using three DNN benchmarks. We evaluate these savings using our custom hardware design accelerator as well as TX1, an embedded GPU platform. Furthermore, our method requires much less memory and silicon real-estate compared to previous dynamic techniques. 

\section{Acknowledgments}
\noindent This work is supported by NSF grant 1420864. We would like to thank NVIDIA Corporation for their generous GPU donation. We also thank Professor Pedro Felzenszwalb for the discussions and his helpful inputs.

\setlength{\bibsep}{2pt plus 0.3ex}
\footnotesize
\bibliographystyle{abbrv}
\bibliographystyle{unsrtnat}
\scriptsize
\bibliography{main.bbl}

\end{document}